\documentclass[conference,letterpaper]{IEEEtran}
\IEEEoverridecommandlockouts
\usepackage[letterpaper, left=1in, right=1in, bottom=1in, top=1in]{geometry}
\usepackage[T1]{fontenc}
\usepackage{amsmath,amssymb,amsfonts}
\usepackage{algorithmic}
\usepackage{graphicx}
\usepackage{textcomp}
\usepackage{xcolor}
\usepackage{multirow}
\usepackage{listings}
\usepackage{subfig}

\usepackage{float}
\usepackage[hyphens]{url}
\usepackage{cleveref}
\usepackage{balance}
\usepackage{makecell}
\usepackage{color}  
\definecolor{webgreen}{rgb}{0,0.5,0} 
\definecolor{webbrown}{rgb}{0.6,0,0} 
\usepackage[style=trad-abbrv,backend=biber,sorting=none,doi=false,url=false,eprint=false,isbn=false]{biblatex}
\def\BibTeX{{\rm B\kern-.05em{\sc i\kern-.025em b}\kern-.08em
    T\kern-.1667em\lower.7ex\hbox{E}\kern-.125emX}}
\begin{document}

\DeclareRobustCommand*{\IEEEauthorrefmark}[1]{%
  \raisebox{0pt}[0pt][0pt]{\textsuperscript{\footnotesize #1}}%
}

\title{An Experiment on Feature Selection using Logistic Regression
\\
}


\author{\IEEEauthorblockN{Raisa Islam\IEEEauthorrefmark{1}, Subhasish Mazumdar\IEEEauthorrefmark{2}, and Rakibul Islam\IEEEauthorrefmark{3}
}
\IEEEauthorblockA{\textit{Dept of Computer Science \& Engineering}\\
\textit{New Mexico Institute of Mining and Technology}\\
Socorro, NM 87801 USA\\
\IEEEauthorrefmark{1}raisa.islam@student.nmt.edu, \IEEEauthorrefmark{2}subhasish.mazumdar@nmt.edu,
\IEEEauthorrefmark{3}mdrakibul.islam@student.nmt.edu}
}



\maketitle
\thispagestyle{plain}
\pagestyle{plain}

\begin{abstract}

In supervised machine learning, feature selection plays a very important role by potentially enhancing explainability and performance as measured by computing time and accuracy-related metrics. In this paper, we investigate a method for feature selection based on the well-known L1 and L2 regularization strategies associated with logistic regression (LR). It is well known that the learned coefficients, which serve as weights, can be used to rank the features.  Our approach is to synthesize the findings of L1 and L2 regularization. For our experiment, we chose the CIC-IDS2018 dataset~\cite{Sharafaldin2018TowardGA} owing partly to its size and also to the existence of two problematic classes that are  hard to separate.  We report first with the exclusion of one of them and then with its inclusion. We ranked features first with L1 and then with L2, and then compared logistic regression with L1 (LR+L1) against that with L2 (LR+L2) by varying the sizes of the feature sets for each of the two rankings. We found no significant difference in accuracy between the two methods once the feature set is selected. 
We chose a synthesis, i.e., only those features that were present in both the sets obtained from L1 and that from L2, and experimented with it on
more complex models like Decision Tree and Random Forest and observed that the accuracy was very close in spite of the small size of the feature set.
Additionally, we also report on the
standard metrics: accuracy, precision, recall, and f1-score.


\end{abstract}


\section{Introduction}
\label{sec:intro}



Feature selection is an effective and efficient data pre-processing method in machine learning (ML) employed to reduce the dimensionality of the data, which can improve the performance of machine learning models and make them more interpretable. Moreover, it has the potential of providing more accurate predictions by removing noise in the data arising from irrelevant features. 
The challenge  is to select a subset of the interesting data features that are most relevant and able to correctly differentiate samples from different classes.





A major problem in supervised ML is overfitting, i.e.,
the situation where the model learns the dataset ``too well",
resulting in a high training data accuracy but a poor test data accuracy during cross-validation. Regularization is one of the most popular methods used to avoid overfitting; it adds a penalty term to the loss function of the ML model that reflects the complexity of the model. L1 regularization tends to assign the coefficients or weights of less important features to zero, thus producing a sparse vector and equivalently a smaller set of good-enough features. On the other hand, L2 regularization tends to shrink the coefficients more uniformly; thus, a sparse vector is not realizable. 

In our proposed method, we attempt to synthesize the two approaches. First, we obtain a ranked ordering of features using logistic regression with L1 regularization, then 
another ordering using L2.
We choose those features that are present in both of these sets
and experimentally test the performance of selected ML models using our set.

Our experiment is on one large dataset which is a non-trivial real-world dataset that is large in volume. Also, it is challenging because there is a problematic class that is hard to separate from another.

Regarding supervised ML models, we targeted the Decision Tree because it is explainable. Since we are studying accuracy, a metric that is not always the highest for these classifiers, we also chose the Random Forest classifier, which is related to the Decision Tree, exhibits excellent accuracy, but sacrifices explainability.
We have found that using our method on Decision Trees and Random Forest, there is a loss of only 0.8 and 0.6\% mean accuracy while reducing the 
feature size by 72\%,
and that regardless of the inclusion of the problematic class.




%


The rest of the paper is organised as follows. In section~\ref{sec:related_work}, we present some background material and prior feature selection techniques. In section~\ref{sec:approach}, we describe the experiment setup and procedure; and in section~\ref{sec:result} present the results and analysis of this case study. Finally, in section~\ref{sec:conclusion}, we offer concluding remarks along  with future research directions.

\section{Background} 
\label{sec:related_work}


\subsection{Logistic regression and Regularization}
Logistic regression (LR) is one of the most popular supervised ML algorithms used for solving the classification problems. Regression analysis models uses coefficients to estimate the features. If the estimates can be narrowed towards zero, then the impact of insignificant features might be reduced. L1 and/or L2 regularization is used with LR to prevent overfitting by adding a penalty term to the cost function. In the context of feature selection, L1 regularization is known to perform feature selection by shrinking the coefficients of less important features to zero which will make some features obsolete. This results in a sparse model where only a subset of features are used. On the other hand, L2 regularization shrinks the coefficients of less important features but does not set them to zero. This results in a model where all features are used, but the less important features have smaller coefficients.

\begin{itemize}
    \item \textit{L1 Regularization} adds a penalty term to the loss function
    equal
    to the sum of the absolute values of the coefficients. The cost function with L1 penalty is given by 

        $$\sum\limits_{i=1}^{n} (y_i - \sum\limits_{j=1}^{m}  x_{ij}.W_j)^2 + \lambda\sum\limits_{j=1}^{m} \vert W_j\vert $$  
    
    \item \textit{L2 Regularization} adds a penalty term to the loss function
    equal to the sum of the squares 
    of the coefficients. Thus cost function with L2 penalty becomes

        $$ \sum\limits_{i=1}^{n} (y_i - \sum\limits_{j=1}^{m}  x_{ij}.W_j)^2 + \lambda\sum\limits_{j=1}^{m}{W_j}^2 $$  
\end{itemize}
Here, $W_j$ is the weight/coefficient of the feature $x_j$ and $\lambda$ is known as the \textit{regularization parameter}.
The terms
$\lambda\sum\limits_{j=1}^{m} \vert W_j\vert$ and $\lambda\sum\limits_{j=1}^{m}{W_j}^2$ are called \textit{penalty terms} for L1 and L2 respectively.



\subsection{Random Forest}
Random Forest (RF)~\cite{breiman2001random} is a \textit{divide-and-conquer} based supervised learning algorithm that can scale the volume while maintaining statistical efficiency of information. It can be applied comprehensively to any prediction problem with few calibrated parameters. RF is a collection of tree predictors that allows the ensemble of trees to choose the most popular class in order to improve classification accuracy. The ensemble generates random vectors to regulate tree growth. Each tree is constructed using random observations from the original dataset.
\subsection{Decision Tree}
A Decision Tree (DT) is a tree-based technique which learns a model by generating the same labeling for the provided data. Similar to RF, DT also uses a \textit{divide-and-conquer} strategy to find the best split points within a tree by employing a greedy search. The main idea is to continually split the dataset into yes/no questions using its features, until each data point is recognized as belonging to a particular class. 

\subsection{Performance evaluation parameters}
The most common way to analyze the performance of any classification model is to use the confusion matrix~\cite{9016261}. Confusion matrix consists of actual label vs predicted labels which helps to visualize the distribution of each class along with a breakdown of error categories. Performance evaluation parameters i.e., accuracy, precision, recall, F1-score, etc., can be computed using the components of confusion matrix.

\paragraph{Accuracy} It is the ratio of the number of accurately predicted instances to the number of total instances. Despite being the most common performance measure, it is not an useful metric for unevenly distributed data.

\paragraph{Precision} The precision is an indicator measuring the exactness of each class. It is the ratio of the number of positive instances that were predicted accurately to the total number of instances that were predicted positive. When a high cost is associated with false positive, precision is a preferable metric.

\paragraph{Recall} The recall is an indicator measuring the completeness of each class. It is the ratio of the number of positive instances that were predicted accurately to the number of actually positive instances. Recall is a preferable metric when the associated cost of false negative is high.

\paragraph{F1-score} F1-score is a function of precision and recall which represents the harmonic mean. F1-score is more appropriate measure for uneven classification problems where a balance between precision and recall is needed. 
\section{Experiment}
\label{sec:approach}

\subsection{Dataset and pre-processing}
\label{sec:pre-processing}




\subsubsection{Dataset}
The CIC-IDS2018 dataset represents observations
gathered over a span of 10 days of network traffic; it
is 
huge in volume representing 16,233,002 samples and is spread over  
ten CSV files~\cite{dataset}. 
For each sample, it gives values for 79 features 
along with one of
15 different target classes of which it is known to be a member. 
Table~\ref{tab:target_classes} lists their names and sample sizes.
%
%

Among the 15 classes, 
the number of samples for \textit{DDOS Attack LOIC UDP }
is small: 1730,
while those for
\textit{Brute Force XSS, 
Brute Force Web}, 
and 
\textit{SQL Injection} 
are extremely small: 230, 611, and 87 respectively.
%
Most real-world datasets are class imbalanced, i.e.,
the number of samples of some classes are grossly different from that of others.
This is a key issue since most ML algorithms assume that 
data is evenly distributed across
classes.
The domination of majority classes over minority ones 
can make ML classifiers biased towards the former
resulting in misclassification of the former.
%
%

\subsubsection{pre-processing}
As a part of a cleaning phase, 
observations with feature values \textit{Infnity}, \textit{NaN} (i.e., a missing value) were dropped. 
Also, there were 59 entries for which the class entry was \textit{Label}, i.e., unknown; these too were dropped.
Finally, we chose not to use the timestamp feature since it is not relevant to the classification \cite{10.1007/978-3-030-44038-1_63}; this left
us with 78 features.
%

We attempted to extract 5,000 random samples from each class that survived
the cleaning phase. 
To avoid drastic class imbalance, we  excluded the three classes with very small sample size: \textit{Brute Force XSS, Brute Force Web}, and 
\textit{SQL Injection},
leaving us with 12 classes. 
We had 5,000 samples each from eleven of them but only 1,730 from the last
(\textit{DDOS Attack LOIC UDP});
thus, we had 56,730 samples.

Moreover,
\textit{DoS attacks-SlowHTTPTest} and 
\textit{FTP-BruteForce} classes are very hard to separate.
We identified 
\textit{DoS attacks-SlowHTTPTest} 
as a \textit{problematic} class. To handle it, we 
split our experiment into two parts, 
excluding the problematic class
in the first part of the experiment and including it in the second. Thus, for the first part, the size of the dataset became 51,730 from 11 classes.


\begin{figure*}[htbp]
\centering
\subfloat[Subfigure 1 list of figures text][Mean accuracy vs top features from L1-rank ordering.]{
\includegraphics[width=0.45\textwidth,height=2in]{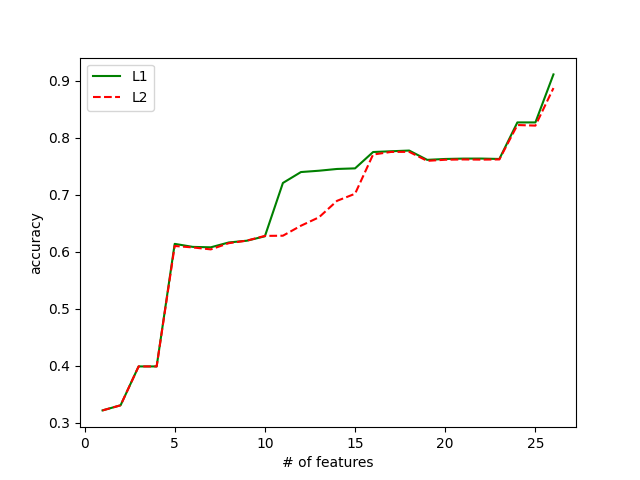}
\label{fig:l1vsl2_5k_a}}
\qquad 
\subfloat[Subfigure 2 list of figures text][Mean accuracy vs top features from L2-rank ordering.]{
\includegraphics[width=0.45\textwidth,height=2in]{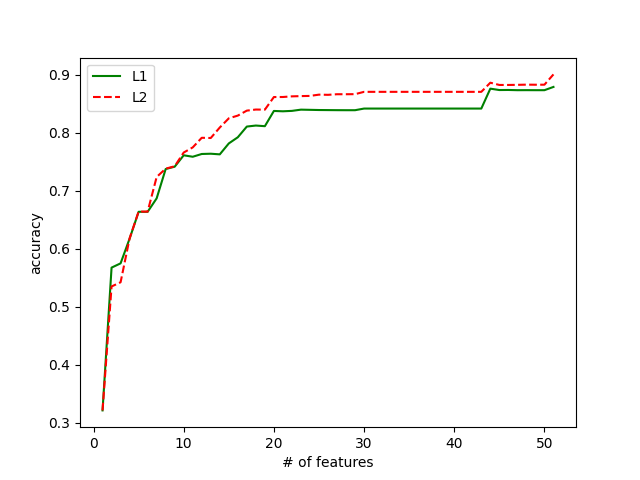}
\label{fig:l1vsl2_5k_b}}
\caption{Accuracy vs number of top features from L1 and L2 orderings.}
\label{fig:l1vsl2_5k}
\end{figure*}

\begin{figure}[!htbp]
    \centering
    \includegraphics[width=0.45\textwidth,height=2in]{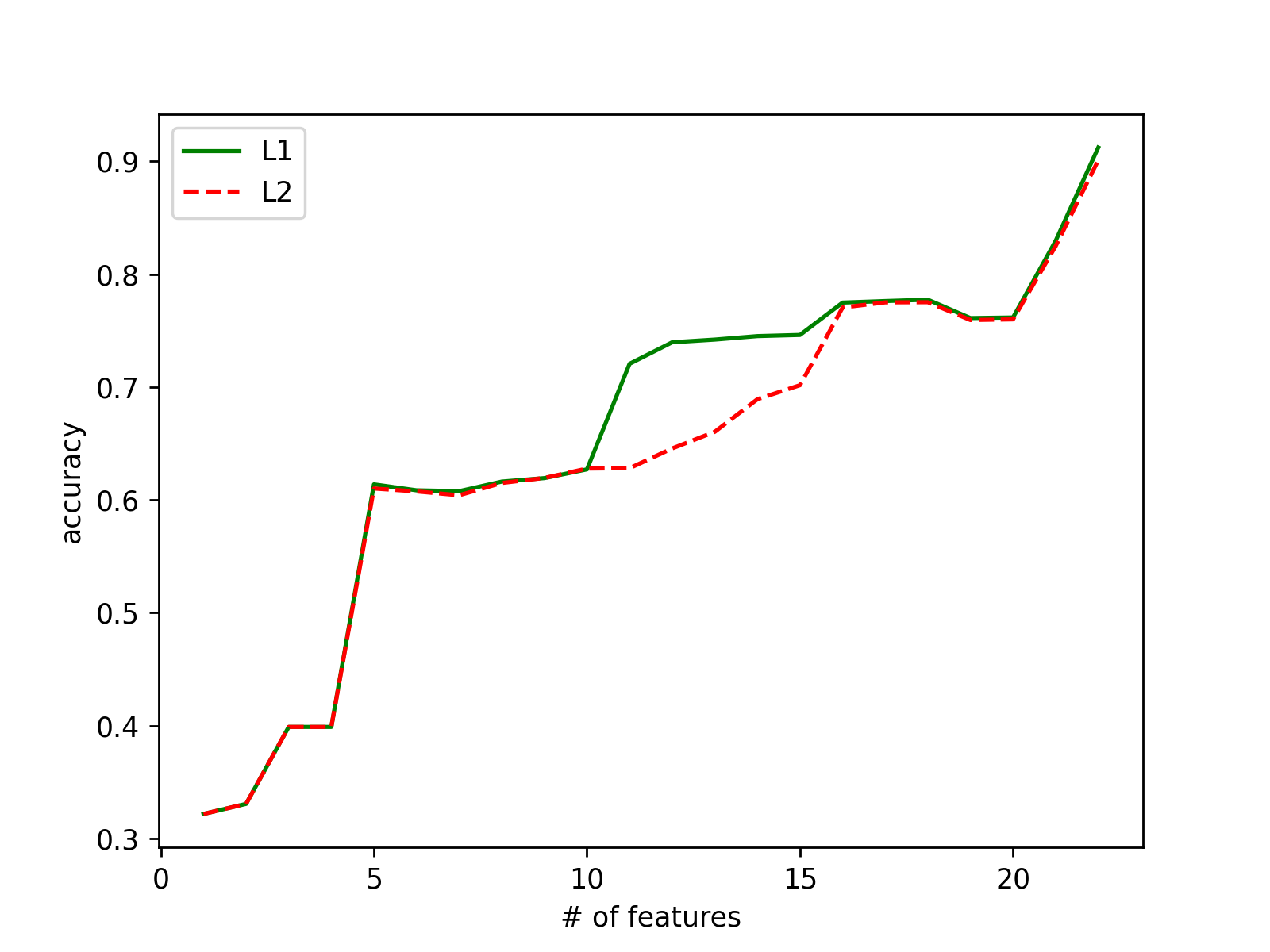}
    \caption{Accuracy vs number of top common features ordered as per Table~\ref{tab:common_features}.}
    \label{fig:common}
\end{figure}


\begin{table}[htb]
    \centering
    \begin{tabular}{|l|l|}
        \hline
        target class & record count \\ \hline \hline
        Benign &  13484708\\ \hline
        DDOS attack-HOIC & 686012\\ \hline
        DDoS attacks-LOIC-HTTP & 576191\\ \hline
        DoS attacks-Hulk & 461912\\ \hline
        Bot & 286191\\ \hline
         FTP-BruteForce &  193360\\ \hline
        SSH-Bruteforce &  187589\\ \hline
        Infilteration &  161934\\ \hline
        DoS attacks-SlowHTTPTest & 139890\\ \hline
        DoS attacks-GoldenEye & 41508\\ \hline
        DoS attacks-Slowloris & 10990\\ \hline
        DDOS Attack LOIC UDP & 1730\\ \hline
        Brute Force Web & 611\\ \hline
        Brute Force-XSS & 230\\ \hline
        SQL Injection & 87\\ \hline
    \end{tabular}
    \caption{target classes}
    \label{tab:target_classes}
\end{table}


\subsection{Experimental setup}
\label{sec:model}

Our experiments were implemented in Python 3 using the scikit learn
({\sf sklearn}) package. This allowed us to execute standard code for training  the machine learning models of interest, as well as for testing and obtaining results; we used {\sf matplotlib} for plotting graphs.


We used a 70 : 30 ratio for training data : test data; consequently,
 in the first part of the experiment,
 the training dataset had 36,211 observations and testing set had 15,519 observations. 
%
%
For the second part of the experiment, we 
added 5000 random samples
of the problematic class \textit{DoS attacks-SlowHTTPTest}.
The 70:30 ratio for training data : test data remained unchanged
leading to  39,711 training samples and 17,019 test ones.
The various class sizes are shown 
in Table~\ref{tab:class_names};
the numbers in the first 11 rows correspond to the first part of the experiment; the last row was obtained from the second part (the changed
values for the other 11 classes in that part are not shown).

\begin{table}[!htbp]
    \centering
    \resizebox{0.97\columnwidth}{!}{%
    \begin{tabular}{|l|l|l|l|l|}
        \hline
        target class & notation & sample size & training set & test set \\ \hline\hline
         Benign & cls-1 & 5000 & 3531 & 1469 \\ \hline
         Bot & cls-2 & 5000 & 3450 & 1550\\ \hline
         DDOS attack-HOIC & cls-3 & 5000 & 3518 & 1482\\ \hline
         DDOS attack-LOIC-UDP & cls-4 & 1730 & 1212 & 518\\ \hline
         DDoS attacks-LOIC-HTTP & cls-5  & 5000 & 3476 & 1524\\ \hline
         DoS attacks-GoldenEye & cls-6 & 5000 & 3442 & 1558\\ \hline
         DoS attacks-Hulk & cls-7  & 5000 & 3545 & 1455\\ \hline
         DoS attacks-Slowloris & cls-8 & 5000 & 3491 & 1509\\ \hline
         FTP-BruteForce & cls-9  & 5000 & 3570 & 1430\\ \hline
         Infilteration & cls-10 & 5000 & 3501 & 1499\\ \hline
         SSH-Bruteforce & cls-11 & 5000 & 3475 & 1525\\ \hline
         DoS attacks-SlowHTTPTest & cls-prb & 5000 & 3518 & 1482\\ \hline
    \end{tabular}
    }
    \caption{Sample dataset description}
    \label{tab:class_names}
\end{table}


As noted earlier, we are interested in Logistic Regression. 
We used the \textit{SAGA} solver 
(a version of SAG, which is stochastic average gradient)
with an inverse regularization parameter equal to 0.5. 

Next, we trained it on the dataset prepared for the first part of the experiment with all the features; this generated coefficient values of each feature and mean accuracy of the model. 

\begin{table}[!htbp]
    \centering
    \resizebox{\columnwidth}{!}{%
    \begin{tabular}{|c|c|l||c|c|l|}
        \hline
        \makecell{L1\\rank} & \makecell{L2\\rank} & feature name & \makecell{L1\\rank} & \makecell{L2\\rank} & feature name\\ \hline \hline
         1 & 1 & Fwd Seg Size Min & 12 & 15 & Init Bwd Win Byts\\ \hline
         2 & 3 & URG Flag Cnt & 13 & 16 & Fwd IAT Max\\ \hline
         3 & 5 & SYN Flag Cnt & 14 & 17 & Bwd IAT Std\\ \hline
         4 & 6 & Fwd PSH Flags & 15 & 18 & Flow IAT Max\\ \hline
         5 & 10 & Bwd IAT Mean & 16 & 12 & Flow Byts/s\\ \hline
         6 & 9 & Bwd IAT Min & 17 & 27 & Flow IAT Mean\\ \hline
         7 & 4 & Fwd Pkt Len Min & 18 & 45 & Protocol\\ \hline
         8 & 21 & Fwd IAT Min & 19 & 2 & Pkt Len Std\\ \hline
         9 & 23 & Flow IAT Std & 22 & 13 & FIN Flag Cnt\\ \hline
         10 & 24 & Fwd IAT Mean & 25 & 51 & RST Flag Cnt\\ \hline
         11 & 22 & Bwd IAT Max & 26 & 44 & PSH Flag Cnt\\
         \hline
    \end{tabular}
    }
    \caption{common features}
    \label{tab:common_features}
\end{table}

As outlined in the Introduction, we are interested in the coefficients learned from logistic regression using L1 and L2 regularization.
Towards that end, we first applied logistic regression with L1 regularization (LR+L1) and sorted the learned coefficients to get an L1-rank ordering of features. 
Similarly, we trained an LR+L2 model and obtained an L2-rank ordering of features.
Next, we sorted both sets of features 
in decreasing order of coefficient value.

\subsubsection{Three experiments: Accuracy vs number of features} 
\label{expt:acc-vs-l1f-l2f}
\mbox{}\\
(a) We tested the learned LR+L1 logistic regression model repeatedly
with an increasing number of features 
form the L1-rank ordering: the top feature, the top two, the top three,
etc., and observed the
improvement in accuracy with increase in the number of features.
This gave us \textbf{MAX\_L1}, the number of features at which the accuracy
reached approximately 95\% of the maximum accuracy (the one obtained using
all features). We took this accuracy level as the baseline value
for the next component.

To find out how different the LR+L2 model would be,
we tested it for the same sequence of features, 
i.e., observed the accuracy of the learned LR+L2 model 
with an increasing number 
of top features from the same L1-rank ordering.

(b) Next, we repeated the above using the L2-rank ordering.
Using the learned LR+L2 model, we first obtained \textbf{MAX\_L2},
the number of features at which the accuracy equaled 
the baseline value obtained earlier
and then observed the improvement of accuracy with increasing features,
and found out how different the LR+L1 model would be for
the same feature sets.


Finally, 
we attempted to synthesize the findings of LR+L1 and LR+L2
by computing the intersection of the L1-rank ordering and
the L2-rank ordering obtaining 
a set of features we call \textit{common features}.

(c) We then re-did the accuracy versus number of 
features experiment using these common features
ordered as per Table~\ref{tab:common_features}
(taking the first eleven from the left column and
the last eleven from the right).

\subsubsection{Twelve experiments}\label{expt:twelve} 
We then conducted twelve experiments which we have given names listed in
Table~\ref{tab:expt-names}. Each name consists of a ML model 
(one of LR+L1, LR+L2, RF for Random Forest, or DT for Decision Tree)
followed by a letter (one of \textit{A, B,} or \textit{C}). The letters
denote the feature set used:
\textit{A} denotes the top MAX\_L1 features from the L1-rank ordering;
\textit{B} the MAX\_L2 features from the L2-rank ordering; and
\textit{C} all the common features.

For example, in \textit{Experiment LR+L1-A}, we 
ran the LR+L1 model using features
ranked 1 through MAX\_L1 in the L1-rank ordering.
We noted the mean accuracy, 
the corresponding confusion matrix, as well as
precision, recall, and F1-score metrics.

As indicated in the last two columns,
we experimented similarly with Random Forest (RF) 
and Decision Tree (DT) classifiers.

\begin{table}[htb]
    \centering
    \begin{tabular}{ ||p{2cm}||c|c|c|c|  }
     \hline
      & \multicolumn{4}{|c|}{ML Models} \\
     \cline{2-5}
      \raisebox{1.5ex}[0cm][0cm]{Features} & LR+L1 & LR+L2 & RF & DT\\ \hline\hline
     L1-rank ordering [$1...MAX\_L1$] & \textit{LR+L1-A} & \textit{LR+L2-A} & \textit{RF-A} &\textit{DT-A}\\\hline
     L2-rank ordering [$1...MAX\_L2$] & \textit{LR+L1-B} & \textit{LR+L2-B} & \textit{RF-B} & \textit{DT-B}\\\hline
     common features (all) & \textit{LR+L1-C} & \textit{LR+L2-C} & \textit{RF-C} & \textit{DT-C}\\
     \hline
    \end{tabular}
    \caption{Experiment names.}
    \label{tab:expt-names}
\end{table}

\begin{figure}[!htbp]
\centering
\subfloat[Subfigure 1][LR+L1-A]{
\includegraphics[width=0.85\columnwidth]{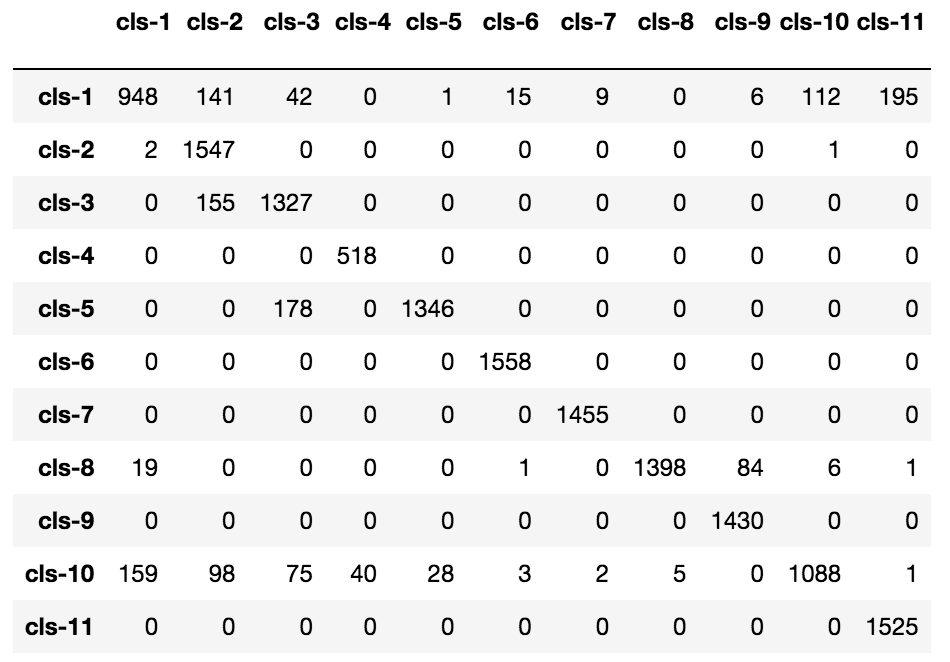}
\label{fig:l1-a}}

\subfloat[Subfigure 2][LR+L1-B]{
\includegraphics[width=0.85\columnwidth]{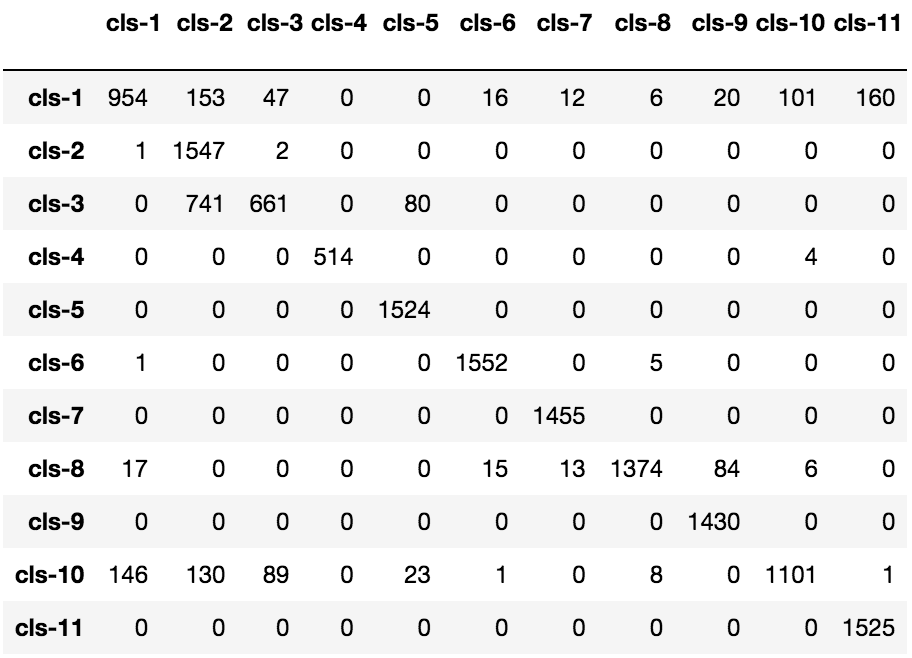}
\label{fig:l1-b}}

\subfloat[Subfigure 3][LR+L1-C]{
\includegraphics[width=0.85\columnwidth]{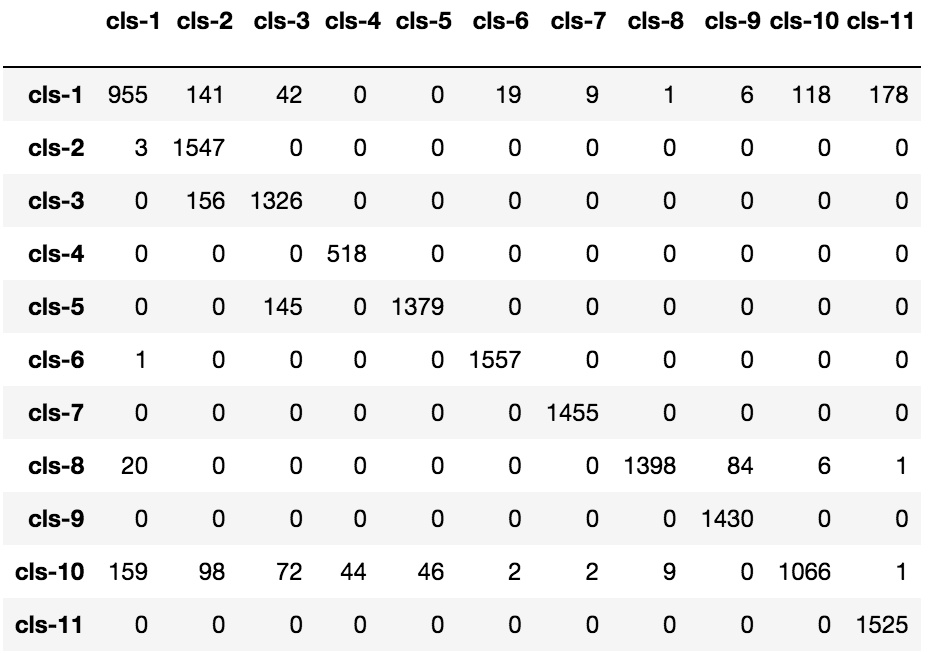}
\label{fig:l1-c}}

\caption{Confusion matrix of LR+L1 model.}
\label{fig:L1-conf}
\end{figure}

\begin{figure}[!htbp]
\centering
\subfloat[Subfigure 1][LR+L2-A]{
\includegraphics[width=0.85\columnwidth]{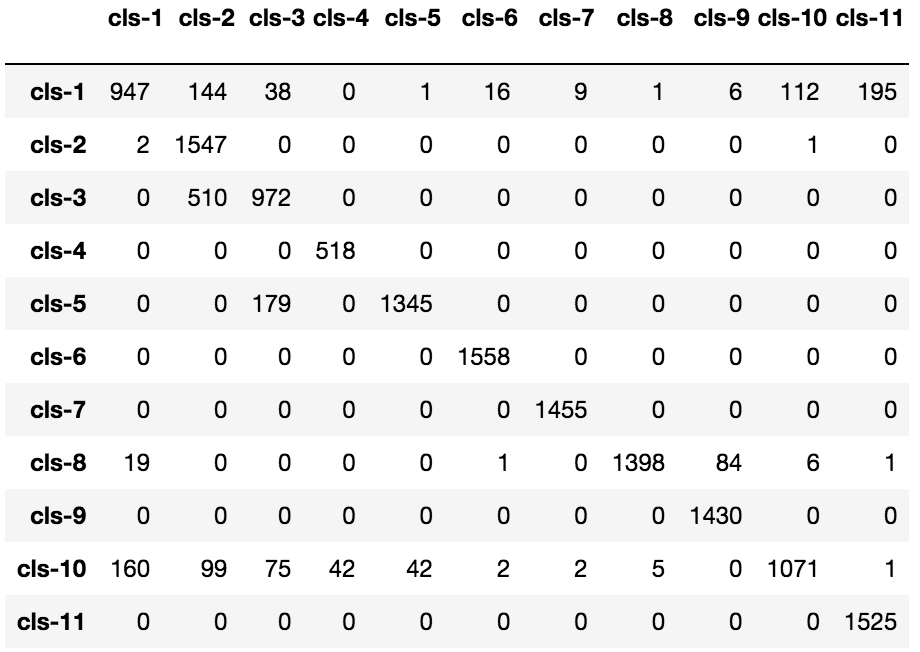}
\label{fig:l2-a}}

\subfloat[Subfigure 2][LR+L2-B]{
\includegraphics[width=0.85\columnwidth]{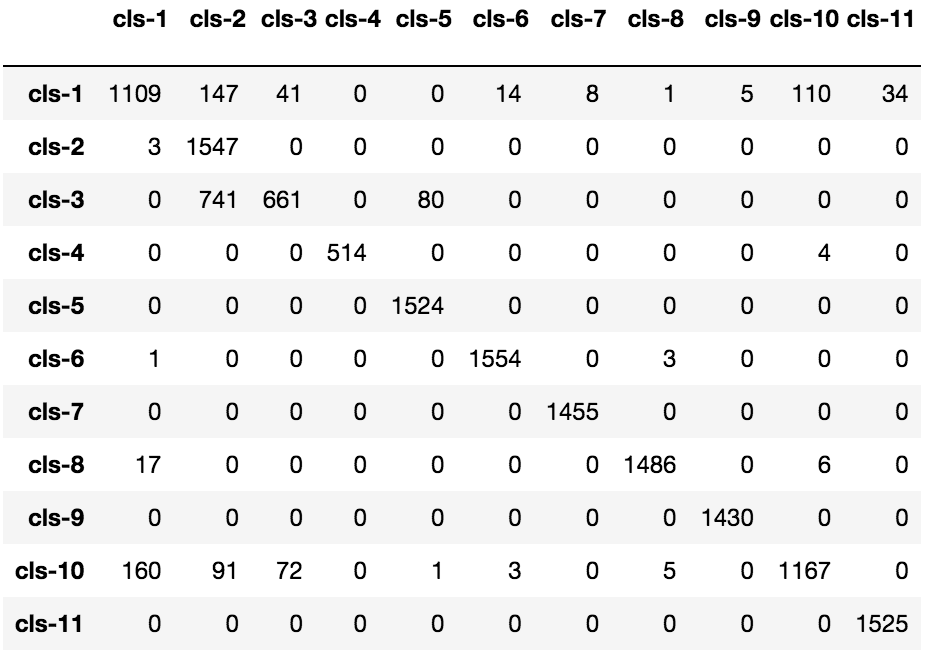}
\label{fig:l2-b}}

\subfloat[Subfigure 3][DT-C]{
\includegraphics[width=0.85\columnwidth]{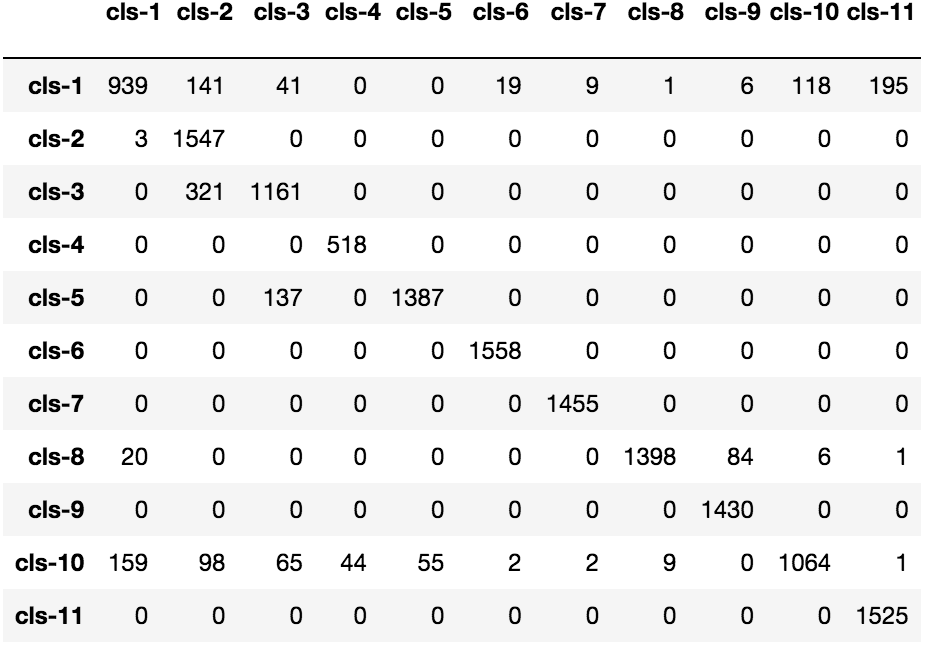}
\label{fig:l2-c}}

\caption{Confusion matrix of LR+L2 model.}
\label{fig:L2-conf}
\end{figure}

\begin{figure}[!htbp]
\centering
\subfloat[Subfigure 1][RF-A]{
\includegraphics[width=0.85\columnwidth]{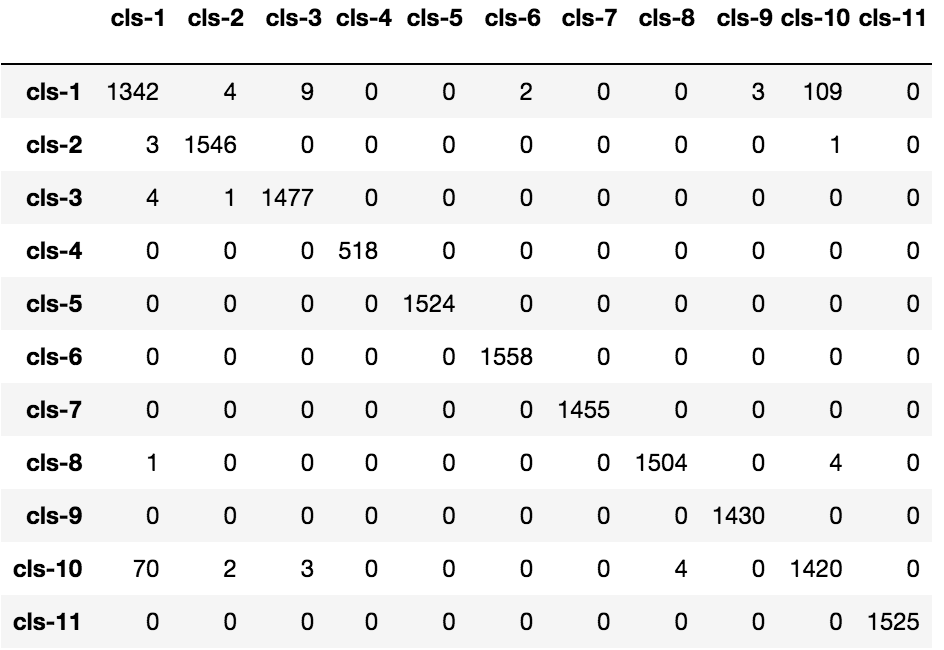}
\label{fig:rf-a}}

\subfloat[Subfigure 2][RF-B]{
\includegraphics[width=0.85\columnwidth]{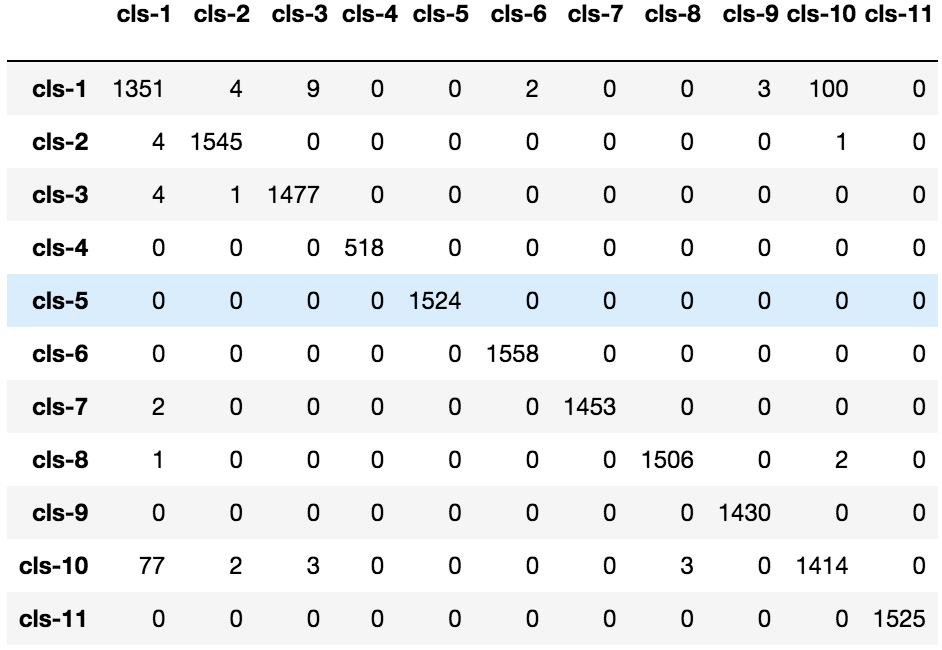}
\label{fig:rf-b}}

\subfloat[Subfigure 3][RF-C]{
\includegraphics[width=0.85\columnwidth]{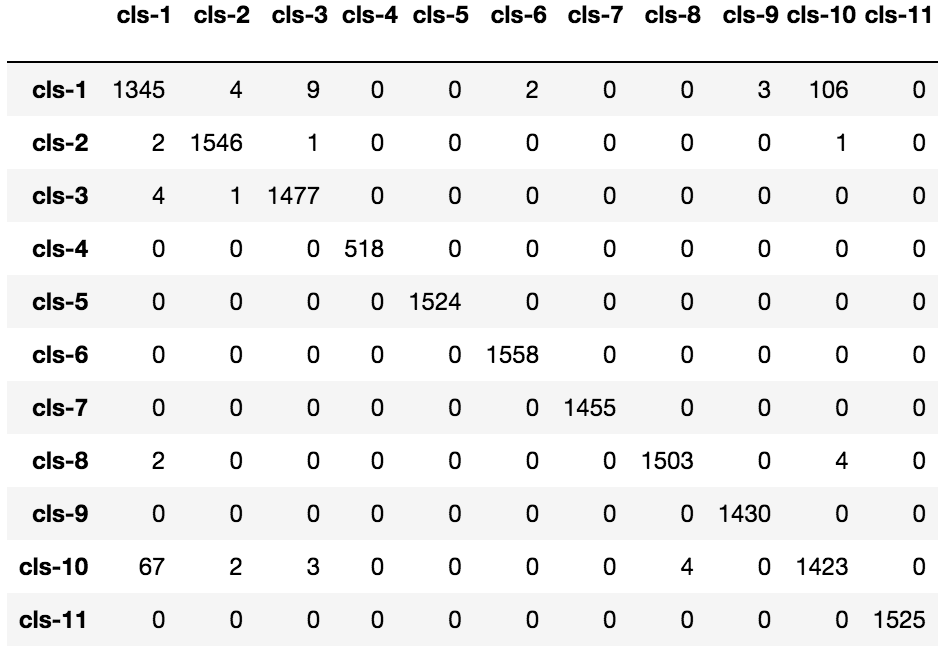}
\label{fig:rf-c}}

\caption{Confusion matrix of RF model.}
\label{fig:RF-conf}
\end{figure}

\begin{figure}[!htbp]
\centering
\subfloat[Subfigure 1][DT-A]{
\includegraphics[width=0.85\columnwidth]{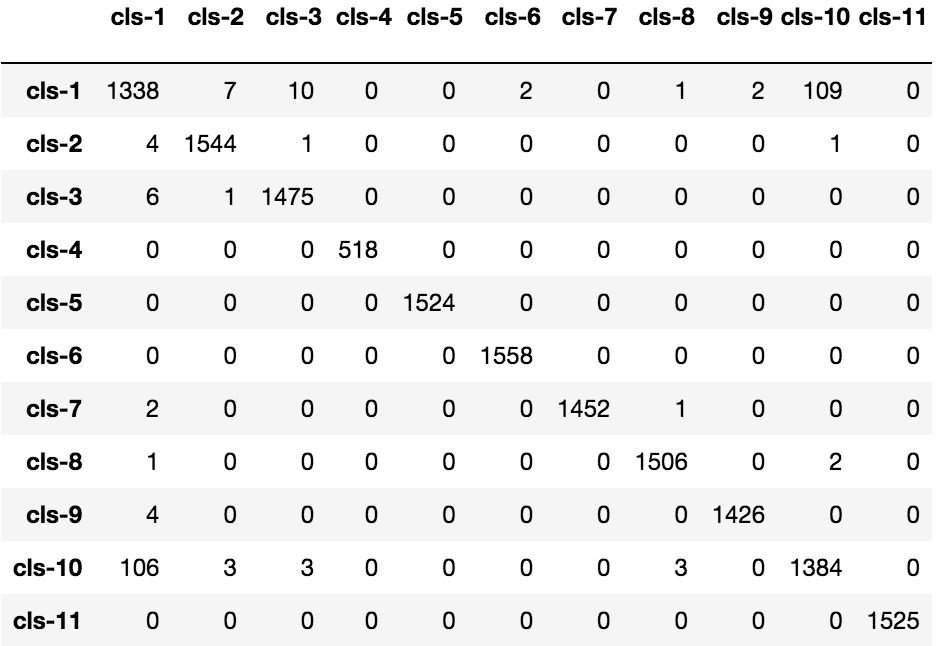}
\label{fig:dt-a}}

\subfloat[Subfigure 2][DT-B]{
\includegraphics[width=0.85\columnwidth]{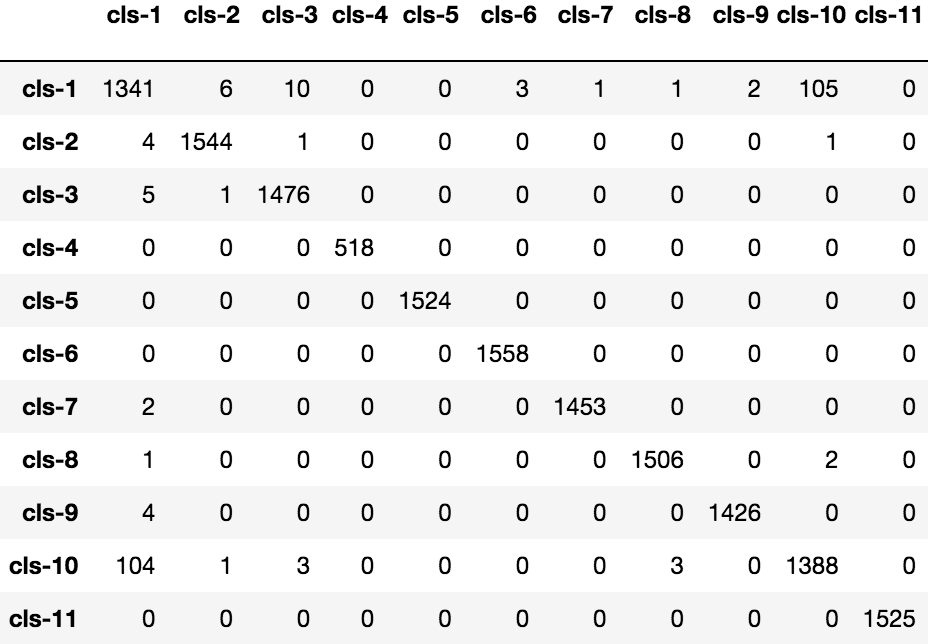}
\label{fig:dt-b}}

\subfloat[Subfigure 3][DT-C]{
\includegraphics[width=0.85\columnwidth]{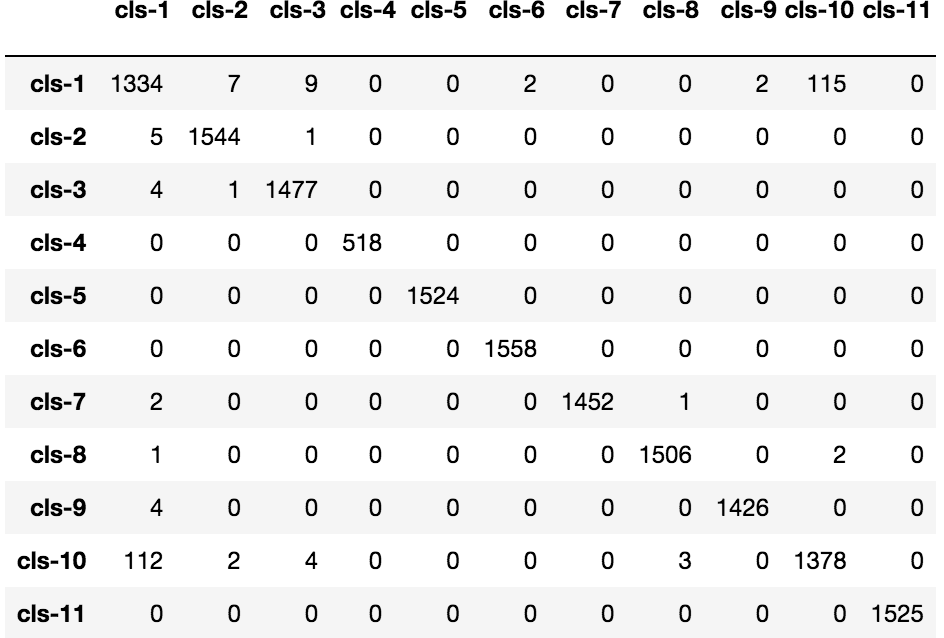}
\label{fig:dt-c}}

\caption{Confusion matrix of DT model.}
\label{fig:DT-conf}
\end{figure}

\section{Result and Analysis}
\label{sec:result}

The common features are listed in Table~\ref{tab:common_features} along with their ranks
in both L1-rank and L2-rank orderings.

The number of common features was 22
drawn from ranks 1 through 26 with average rank of 12
from the L1-rank and
from ranks 1 through 51 with average rank of 18
from the L2-rank. 
This showed that the common features primarily follow
the L1-rank ordering.

\subsection{First part:  excluding the problematic class}
As mentioned earlier, the first part of the experiment
excludes the problematic class.

\subsubsection{Accuracy vs number of features}
We described three experiments in 
Subsection~\ref{expt:acc-vs-l1f-l2f}. 
The results of the first, i.e., (a) are displayed in Fig~\ref{fig:l1vsl2_5k_a}. We learned that MAX\_L1 = 26, i.e.,
with the 26 top L1-ranked features, we obtained 
a 91.11\% accuracy, which was approximately 95\% of the
maximum  value of 95.69\% that we found possible with logistic regression with L1 regularization using all features.

Furthermore, we observed that
the performances of LR with L1 and L2 are very similar
(the divergence around 11 -- 16 features can be attributed
to the ordering of features being L1-ranked).

Similarly, the results of the second part, 
(outlined in Subsection~\ref{expt:acc-vs-l1f-l2f}(b))
are displayed in 
Fig.~\ref{fig:l1vsl2_5k_b}.
Here, we found MAX\_L2 = 51, i.e.,
with the 51 top L2-ranked features, we obtained 
a 90.03\% accuracy, on par with the LR+L1 observations.

As in part (a), we observed similar performances
of LR with L1 and L2 (the divergence can be similarly
attributed to the use of the L2-ranked ordering).

\begin{table*}[htb]
    \centering
    \begin{tabular}{|l|c|c|c|c|c|c|}
        \hline
          & all 78 features & \textit{A} (L1 features) &
          \textit{B} (L2 features) &
          \textit{C} (common features) 
          & top 22 L1 features & top 22 L2 features\\ \hline
         \textit{LR+L1-} 
         & 0.9569 & 0.9111 & 0.8787 & 0.9122 & 0.7634 & 0.8375\\ \hline
         \textit{LR+L2-} 
         & 0.9689 & 0.887 & 0.9003 & 0.901 & 0.7615 & 0.8625 \\ \hline
         \textit{RF-} 
         & 0.9917 & 0.9858 & 0.9826 & 0.986 & 0.9823 & 0.9826 \\ \hline
         \textit{DT-} 
         & 0.9903 & 0.9827 & 0.9832 & 0.9822 & 0.9796 & 0.9785 \\
         \hline
    \end{tabular}
    \caption{Mean accuracy comparison excluding problematic class.}
    \label{tab:acc_comparison}
\end{table*}

The results of the third part 
(outlined in Subsection~\ref{expt:acc-vs-l1f-l2f}(c))
are displayed in
Fig~\ref{fig:common}.
Here we observed that we 22 features, we obtained
an accuracy of 91.22\% with LR+L1 and 
90.1\% with LR+L2,
both slightly higher than their corresponding counterparts
with 26 and 51 features respectively).
These values are summarized in
Table~\ref{tab:acc_comparison}. 
It is apparent that using common features (which is only 28\% of the original features), the drop in mean accuracy is not significant when compared with the values generated with all 78 features.

%

%

\subsubsection{Twelve experiments}
We described 12 experiments in Subsection~\ref{expt:twelve}.
Their results for mean accuracy are listed 
in Table~\ref{tab:acc_comparison}
under
the columns `L1 features', `L2 features', and 'common features'
corresponding to the suffix \textit{A, B}, and \textit{C} in
the experiment names (see Table~\ref{tab:expt-names}).
In other words, 0.9111 is the accuracy obtained in 
\textit{Experiment LR+L1-A}, etc.

The last two columns provide more observations per row: 
where the experiments using L1-rank features (-A)
and L2-rank features (-B) are restricted to the
top 22 of their features to compare with an equal number
of features from the set of common features.
They show the superiority of the set of common features
does not come from their size.

\begin{table}[!htbp]
    \centering
    \begin{tabular}{|l|c|c|c|c|}
        \hline
         & accuracy & precision & recall & F1-score \\ \hline
         LR+L1 & 0.91 & 0.91 & 0.92 & 0.92\\ \hline
         LR+L2 & 0.90 & 0.91 & 0.91 & 0.90\\ \hline
         RF & 0.99 & 0.99 & 0.99 & 0.99\\ \hline
         DT & 0.98 & 0.98 & 0.98 & 0.98\\ \hline
    \end{tabular}
    \caption{Performance metrics excluding problematic class using common features.}
    \label{tab:perform_comp}
\end{table}

The accuracy along with precision, recall, and F1-scores
are summarized for all four ML models we have used
in Table~\ref{tab:perform_comp}. It is visible that both RF and DT works very well with the selected features; all performance metrics have a score of 98\% or higher. For completeness, we also report the corresponding 
confusion matrices in Fig.~\ref{fig:L1-conf}--\ref{fig:DT-conf}.

\begin{table*}[hbt!]
    \centering
    \begin{tabular}{|l|c|c|c|c|c|c|}
        \hline
          & all 78 features & \textit{A} (L1 features) & \textit{B} (L2 features) & \textit{C} (common features) & top 22 L1 features & top 22 L2 features\\ \hline
         \textit{LR+L1-} & 0.8927 & 0.7983 & 0.8307 & 0.8012 & 0.6965 & 0.8013 \\ \hline
         \textit{LR+L2-} & 0.9115 & 0.7979 & 0.8469 & 0.8019 & 0.6988 & 0.8113 \\ \hline
         \textit{RF-} & 0.9385 & 0.9333 & 0.9328 & 0.933 & 0.9297 & 0.9295 \\ \hline
         \textit{DT-} & 0.9381 & 0.9304 & 0.9307 & 0.9306 & 0.9277 & 0.9263 \\
         \hline
    \end{tabular}
    \caption{Mean accuracy comparison including problematic class.}
    \label{tab:acc_comparison_inc}
\end{table*}

\subsection{Second part: including the problematic class}
For the second part, we performed 12 experiments described in Subsection~\ref{expt:twelve} after including the
problematic class in the dataset (56,730 samples).
The comparison of mean accuracy using different ML models 
for this case
is presented in Table~\ref{tab:acc_comparison_inc}. Comparing Table~\ref{tab:acc_comparison} and Table~\ref{tab:acc_comparison_inc}, it is clear that the mean accuracy dropped drastically for all the models. 
On top of that, none of the feature subsets performed well for 
either  LR+L1 or LR+L2. However, for with RF and DT,
the mean accuracy achieved using the 22 common features 
are very close to that achieved with all 78 features,
showing the effectiveness of the set of common features.

The confusion matrix after inclusion is very similar 
for all the classes except for the problematic class
\textit{DoS attacks-SlowHTTPTest} (cls-prb) and
\textit{FTP-BruteForce} (cls-9); these are the two classes that are
hard to separate. 
We will refer to the latter (cls-9) as the \textit{confounding class}. 

So, instead of showing the entire confusion matrix, we focus on the 
sub-matrix corresponding to these two classes.
Table~\ref{tab:conf_inc} shows the values of that sub-matrix.
Both LR+L1 and LR+L2 using L1-ranked ordering or common features, identified all the samples of confounding class as the problematic class 
(see Table~\ref{tab:l1-a-inc}, Table~\ref{tab:l1-c-inc}, Table~\ref{tab:l2-a-inc}, Table~\ref{tab:l2-c-inc}). 
The situation is better for both RF and DT regardless of the feature set:
both models identified the problematic class correctly 97\% of the time although they correctly identify the confounding class only 42\% of the time. 
It is slightly better for L2-rank ordering features with both LR+L1 and LR+L2 since they identify the problematic class correctly 55\% of the time and 
the confounding class 80\%.

Since these appear to be tradeoffs, we present the values of recall
for both the problematic class and the confounding class in Table~\ref{tab:recall_comparison_slowhttp} and Table~\ref{tab:recall_comparison_ftp_bruteforce}.




\begin{table*}%
  \centering
  \subfloat[LR+L1-A]{
    \begin{tabular}{|c|c|c|}
        \hline
          &  cls-prb & cls-9 \\ \hline
         cls-prb & 1482 & 0\\ \hline
         cls-9 & 1490 & 0\\ \hline
    \end{tabular}
    \label{tab:l1-a-inc}
  }
  \subfloat[LR+L1-B]{
    \begin{tabular}{|c|c|c|}
        \hline
          &  cls-prb & cls-9 \\ \hline
         cls-prb & 812 & 670 \\ \hline
         cls-9 & 316 & 1174 \\ \hline
    \end{tabular}
    \label{tab:l1-b-inc}
  }
  \subfloat[LR+L1-C]{
    \begin{tabular}{|c|c|c|}
        \hline
          & cls-prb & cls-9 \\ \hline
         cls-prb & 1482 & 0\\ \hline
         cls-9 & 1490 & 0\\ \hline
    \end{tabular}
    \label{tab:l1-c-inc}
  }
  \qquad
  \subfloat[LR+L2-A]{
    \begin{tabular}{|c|c|c|}
        \hline
          & cls-prb & cls-9 \\ \hline
         cls-prb & 1482 & 0\\ \hline
         cls-9 & 1490 & 0\\ \hline
    \end{tabular}
    \label{tab:l2-a-inc}
  }
  \subfloat[LR+L2-B]{
    \begin{tabular}{|c|c|c|}
        \hline
          & cls-prb & cls-9 \\ \hline
         cls-prb & 812 & 670 \\ \hline
         cls-9 & 316 & 1174 \\ \hline
    \end{tabular}
    \label{tab:l2-b-inc}
  }
  \subfloat[LR+L2-C]{
    \begin{tabular}{|c|c|c|}
        \hline
          & cls-prb & cls-9 \\ \hline
         cls-prb & 1482 & 0\\ \hline
         cls-9 & 1490 & 0\\ \hline
    \end{tabular}
    \label{tab:l2-c-inc}
  }
  \qquad
  \subfloat[RF-A]{
    \begin{tabular}{|c|c|c|}
        \hline
          & cls-prb & cls-9 \\ \hline
         cls-prb & 1433 & 49\\ \hline
         cls-9 & 859 & 631\\ \hline
    \end{tabular}
    \label{tab:rf-a-inc}
  }
  \subfloat[RF-B]{
    \begin{tabular}{|c|c|c|}
        \hline
          & cls-prb & cls-9 \\ \hline
         cls-prb & 1433 & 49\\ \hline
         cls-9 & 859 & 631\\ \hline
    \end{tabular}
    \label{tab:rf-b-inc}
  }
  \subfloat[RF-C]{
    \begin{tabular}{|c|c|c|}
        \hline
          & cls-prb & cls-9 \\ \hline
         cls-prb & 1435 & 47\\ \hline
         cls-9 & 861 & 629\\ \hline
    \end{tabular}
    \label{tab:rf-c-inc}
  }
  \qquad
  \subfloat[DT-A]{
    \begin{tabular}{|c|c|c|}
        \hline
          & cls-prb & cls-9 \\ \hline
         cls-prb & 1435 & 47\\ \hline
         cls-9 & 861 & 629\\ \hline
    \end{tabular}
    \label{tab:dt-a-inc}
  }
  \subfloat[DT-B]{
    \begin{tabular}{|c|c|c|}
        \hline
          & cls-prb & cls-9 \\ \hline
         cls-prb & 1435 & 47\\ \hline
         cls-9 & 861 & 629\\ \hline
    \end{tabular}
    \label{tab:dt-b-inc}
  }
  \subfloat[DT-C]{
    \begin{tabular}{|c|c|c|}
        \hline
          & cls-prb & cls-9 \\ \hline
         cls-prb & 1435 & 47\\ \hline
         cls-9 & 861 & 629\\ \hline
    \end{tabular}
    \label{tab:dt-c-inc}
  }
  \qquad
  \caption{Confusion (sub-)matrix only for the problematic class and its confounding class. }%
  \label{tab:conf_inc}%
\end{table*}


\begin{table*}[!htbp]
    \centering
    \begin{tabular}{|l|c|c|c|c|c|}
        \hline
          & L1 features & L2 features & common features & top 22 L1 features & top 22 L2 features\\ \hline
         LR+L1 & 1.00 & 0.55 & 1.00 & 1.00 & 0.54\\ \hline
         LR+L2 & 1.00 & 0.55 & 1.00 & 1.00 & 0.55 \\ \hline
         RF & 0.97 & 0.97 & 0.97 & 0.97 & 0.97 \\ \hline
         DT & 0.97 & 0.97 & 0.97 & 0.97 & 0.97 \\
         \hline
    \end{tabular}
    \caption{Comparison of recall values only for the problematic class \textit{DoS attacks-SlowHTTPTest}.}
    \label{tab:recall_comparison_slowhttp}
\end{table*}

\begin{table*}[!htbp]
    \centering
    \begin{tabular}{|l|c|c|c|c|c|}
        \hline
          & L1 features & L2 features & common features & top 22 L1 features & top 22 L2 features\\ \hline
         LR+L1 & 0.00 & 0.79 & 0.00 & 0.00 & 0.79\\ \hline
         LR+L2 & 0.00 & 0.79 & 0.00 & 0.00 & 0.79 \\ \hline
         RF & 0.42 & 0.42 & 0.42 & 0.42 & 0.42 \\ \hline
         DT & 0.42 & 0.42 & 0.42 & 0.42 & 0.42 \\
         \hline
    \end{tabular}
    \caption{Comparison of recall values only for the confounding class \textit{FTP-BruteForce}.}
    \label{tab:recall_comparison_ftp_bruteforce}
\end{table*}

\section{Conclusion \& Future Work}
\label{sec:conclusion}

In this research, we experimented on
a logistic regression based feature selection method to reduce the number of features required to train a supervised ML model. We chose the
CIC-IDS2018 dataset to analyze the performance of feature selection method on both linear and complex machine learning models.
Using our proposed method synthesizing top features from LR+L1 and LR+L2, 
we obtained a small feature set of size 22,
a 72\% reduction from the original.
(22 of 78) 
and observed that
(a) it was the most important set of that size since it performed
better than subsets of that size from either the L1-rank ordering or the
L2-rank ordering for every ML model --- both linear and complex;
(b) we could reach or exceed the baseline accuracy target set 
(see Section~\ref{expt:acc-vs-l1f-l2f})
for LR with less features --- 
15\% less for L1 and 57\% less for L2.
Note that it may appear that the accuracy for our feature set is
5 to 7\% lower compared with full 78 feature set,
but that is not a reasonable conclusion
since we had set a lower baseline accuracy target.
When the problematic class was included,
the mean accuracy dropped owing to that class and one confounding class
as we have shown with the confusion submatrix;
(c) going beyond linear models to Random Forest and Decision Trees,
our feature set showed an accuracy loss of less than 1\%
with a 72\% reduction in feature size. This was valid even when
the problematic class was included.

One limitation this scheme may encounter is a dataset in which
the common set is the same size as the L1-ranked ordering in which
case it would  become equivalent to the latter.
%
%
Also, our experiment was on a single dataset, which is not enough to prove the efficacy of this method in general.

In the future, we will do performance analysis on other datasets and provide a heuristic to predict the resulting number of 
features number.

\balance        
\printbibliography

@ARTICLE{9016261, 
author={Akhter, Muhammad Pervez and Jiangbin, Zheng and Naqvi, Irfan Raza and Abdelmajeed, Mohammed and Mehmood, Atif and Sadiq, Muhammad Tariq}, 
journal={IEEE Access}, 
title={Document-Level Text Classification Using Single-Layer Multisize Filters Convolutional Neural Network}, year={2020},  volume={8},  number={}, pages={42689-42707}, doi={10.1109/ACCESS.2020.2976744}}

@article{breiman2001random, 
title={Random forests}, author={Breiman, Leo}, 
journal={Machine learning}, 
volume={45}, number={1}, pages={5--32}, year={2001}, publisher={Springer} }

@inproceedings{Sharafaldin2018TowardGA, 
title={Toward Generating a New Intrusion Detection Dataset and Intrusion Traffic Characterization}, 
author={Iman Sharafaldin and Arash Habibi Lashkari and Ali A. Ghorbani}, booktitle={ICISSP}, year={2018} }

@inbook{10.1007/978-3-030-44038-1_63,
author = {Catillo, Marta and Rak, Massimiliano and Umberto, Villano},
year = {2020},
month = {03},
pages = {687-696},
title = {2L-ZED-IDS: A Two-Level Anomaly Detector for Multiple Attack Classes},
isbn = {978-3-030-44037-4},
doi = {10.1007/978-3-030-44038-1_63}
}

@misc{dataset, 
    title = {{CIC-IDS2018} dataset}, 
    howpublished = {\url{https://www.unb.ca/cic/datasets/ids-2018.html}} 
}

\end{document}